# Nearest Neighbor Based Out-of-Distribution Detection in Remote Sensing Scene Classification

Dajana Dimitrić and Vladimir Risojević
Faculty of Electrical Engineering
University of Banja Luka
Banja Luka, Bosnia and Herzegovina
jovanicdajana@gmail.com, vladimir.risojevic@etf.unibl.org

Mitar Simić
Faculty of Technical Sciences
University of Novi Sad
Novi Sad, Serbia
mitar.simic@uns.ac.rs

*Abstract*— Deep learning models for image classification are typically trained under the "closed-world" assumption with a predefined set of image classes. However, when the models are deployed they may be faced with input images not belonging to the classes encountered during training. This type of scenario is common in remote sensing image classification where images come from different geographic areas, sensors, and imaging conditions. In this paper we deal with the problem of detecting remote sensing images coming from a different distribution compared to the training data – out of distribution images. We propose a benchmark for out of distribution detection in remote sensing scene classification and evaluate detectors based on maximum softmax probability and nearest neighbors. The experimental results show convincing advantages of the method based on nearest neighbors.

## I. Introduction

Similarly as in other domains relying on visual data analysis, deep learning methods have achieved state-of-the-art results in remote sensing image classification. However, even when a satisfactory classification accuracy on in-distribution (ID) data is achieved, there is a question: What if an input sample does not belong to any of the classes seen during the training phase? How to recognize it and label it as an out-of-distribution (OOD) sample? This type of scenario is common in remote sensing image classification where images come from different geographic areas, sensors, and imaging conditions. Unfortunately, deep learning classifiers are likely to fail in an unexpected way when faced with OOD data. Assuming that the classifier will be deployed in an "open-world" environment, this weakness of standard deep learning classifiers motivates us to examine approaches for OOD image detection. In spite of the importance of this problem, the literature on OOD detection in remote sensing image classification is scarce, with [1] and [2] being notable exceptions.

In [3] a method for OOD detection based on maximum softmax probability (MSP) calculated by a neural network is proposed. It is reported that MSP based approach is effective for OOD image detection compared to more complex methods designed for the purpose of OOD detection. Because of its simplicity, we consider the MSP based approach as a baseline and we examine its performance in remote sensing OOD detection.

Recently, distance-based methods, which declare a sample as OOD if it is far enough from the training data, have shown good results in combination with image features extracted using a deep learning model, [4], [5].

Both of the mentioned methods are based on information that can be extracted from a standard classifier trained on ID data. Since it is very hard to collect representative OOD training dataset, extracting information from a classifier trained under "closed-world" assumptions is a big advantage.

In this paper, we first propose a benchmark for evaluation of OOD detectors in remote sensing image classification and compare a k-nearest neighbors (KNN)-based OOD detector to a baseline detector based on maximum softmax probability. Our experimental results show that the KNN-based detector improves AUROC by 17% and FPR by 70%, compared to the MSP baseline.

## II. Methods

OOD detector is formulated as a binary classifier. Test samples can belong to one of the two classes: (1) positive ID class and (2) negative OOD class, depending on the chosen out of distribution measure and a threshold. The threshold is usually chosen so that 95% of ID data are correctly classified [5], which means that the true positive rate of the ID dataset is equal to 0.95.

Decision whether a test sample is ID or OOD can be made by using softmax probabilities from the network output [3]. For this purpose, we first train the neural network on the ID training set. If the maximum softmax probability of a test sample, $p^{max}$, is greater than threshold, the test sample is ID. Otherwise, the detector recognizes the test image as an OOD sample.

$$p^{max} > \theta \qquad (1)$$

In [4] a method for OOD detection based on the distances of the test image from the training images is proposed. In this method image embeddings, obtained from the layer before the final softmax classifier, are used for computing the distances between images. If the average distance of the test image from $k$ nearest training images is larger than a threshold, the test image is OOD. Formally, if $z_i$ represent the feature vectors of $k$ nearest training samples, and $z^*$ represents the feature vector of the test sample, and $\theta$ represents the chosen threshold, then detector labels a test image as OOD if

$$\frac{1}{k} \cdot \sum_{i=1}^{k} d(z_i, z^*) \geq \theta \quad (2)$$

III. EXPERIMENTS

For the experiments in this paper we use 4 datasets. Our base dataset is NWPU-RESISC45 [6], which contains 31,500 images divided into 45 scene categories. All images are 256x256 pixels with spatial resolutions ranging from 0.2 to 30 m. We consider urban classes: airplane, airport, baseball diamond, basketball court, bridge, church, commercial area, dense residential, freeway, ground track field, harbor, industrial area, intersection, medium residential, mobile home park, overpass, palace, parking lot, railway, railway station, roundabout, runway, ship, stadium, storage tank, tennis court, thermal power station as ID, and rural classes: beach, chaparral, circular farmland, cloud, desert, forest, golf course, island, lake, meadow, mountain, rectangular farmland, river, sea ice, snowberg, sparse residential, terrace, wetland as OOD. In this way, the ID dataset consists of 18,900 images belonging to 27 classes, while the OOD dataset includes 12,600 images from 18 classes. Additionally, we split the ID dataset by taking 500 images from each class for training, 140 for validation, and 60 for test. Overall, our training dataset contains 13,500 images, validation dataset 3,780 images, and test dataset 1,620 images.

As OOD datasets we also use Imagenette [7] and Food5K [8]. Imagenette is a subset of 10 easily classified classes from ImageNet: tench, English springer, cassette player, chain saw, church, French horn, garbage truck, gas pump, golf ball and parachute. For OOD detector testing we use only the validation subset, which contains 3,925 images.

Food5K dataset contains 5,000 food and non-food images divided into 3 sets - training, validation and test. Each set contains food and non-food categories, with 1,500 images in the training, 500 images in the validation and 500 images in the test set. For the OOD detector testing purposes, food images from all 3 sets are used, which amounts to a total of 2,500 images.

Finally, in order to test the OOD detector under a distribution shift, we use PatternNet [9]. PatternNet is a remote sensing dataset, which contains 38 classes with 800 256x256 images in each. The spatial resolutions of images are from 0.062 - 4.693 m. The classes corresponding to the NWPU-RESISC45 ID classes: airplane, baseball field, basketball court, bridge, crosswalk, dense residential, football field, freeway, harbor, intersection, mobile home park, overpass, parking lot, railway, runway, storage tank and tennis court are considered as ID, and remaining classes: beach, cemetery, chaparral, christmas tree farm, closed road, coastal mansion, ferry terminal, forest, golf course, nursing home, oil gas field, oil well, parking space, river, runway marking, shipping yard, solar panel, sparse residential, swimming pool, transformer station and wastewater treatment plant are considered as OOD. The ID dataset consists of 13,600 images from 17 classes and the OOD dataset consists of 16,800 images from 21 classes. It should be noted that we use the ID images from PatternNet only for testing the OOD classifier.

As a classifier we use ResNet50 [10], pre-trained on ImageNet and fine-tuned on the training ID set. The model is adapted by replacing the 1000 neurons from the last layer with only 27, corresponding to the number of classes in the ID dataset. We extract image features before applying the softmax activation function. In this way, the dimensionality of the image embeddings obtained from the model is 2048 and the dimensionality of softmax probabilities is equal to 27. In order to evaluate the impact of the classifier accuracy on the performance of an OOD detector, we implement different variants of network training. In each of the training variants images are converted from RGB to BGR and each color channel is zero-centered. Also, in each of the 4 cases Adam optimizer is used and the network is trained for 50 epochs. In variants 1) and 2) the learning rate is constant and equal to 0.0001, while in the variants 3) and 4) it is variable with the initial value of 0.0001. In summary, the variants for network training are:

1) Base: No data augmentation techniques are used and the learning rate is kept constant.

2) Augmentations: In order to reduce overfitting and improve the performance of deep neural network on ID data, we add augmentation techniques: rotation in range of 90 degrees, zoom in range of 0.15, width and height shift in range of 0.2, shear in range of 0.15, horizontal and vertical flip.

3) Scheduler: To reduce learning rate when learning stagnates, we use ReduceLROnPlateau scheduler with default parameters: validation loss as quantity to be monitored, 0.1 as factor by which the learning rate will be reduced, 10 as number of epochs with no improvement after which learning rate will be reduced and 0 as lower bound on the learning rate.

4) Augmentations and scheduler: Both augmentation and learning rate reducing techniques are added.

For deep learning we used Keras and trained all the networks on Nvidia K80 GPUs available on Google Colab.

Using the trained neural network, we extract features of training images, which we then use for scoring whether a test image is OOD. For scoring we use average cosine distance between the test image and $k$-nearest neighbors from the training set. To find $k$-nearest neighbors, we use the KNeighborsClassifier from the neighbors module of the Scikit-learn library. For comparison of OOD detection methods we used 5 nearest neighbors ($k = 5$). We also examined the influence of the value of $k$ on the detector performance.

For detector performance evaluation and comparison, we use two metrics: (1) false positive rate (FPR) and (2) Area Under the Receiver Operating Characteristic curve (AUROC). To calculate FPR and AUROC values, four measures are used: (1) true positive (TP) - number of ID samples classified as ID, (2) true negative (TN) - number of OOD samples classified as OOD, (3) false positive (FP) - number of OOD samples classified as ID and (4) false negative (FN) - number of ID samples classified as OOD.

False positive rate represents error a detector makes on the OOD dataset, FPR = FP / (FP + TN). It is a threshold-dependent metric and it is calculated when the true positive rate of ID samples, TPR = TP / (TP + FN) is 95%. The lower false positive rate means a more effective detector.

The ROC curve shows true positive rate values against false positive rate values, calculated at different thresholds. AUROC is a threshold-independent metric and a classifier is better if its AUROC is higher.

Choices of thresholds in (1) and (2) represent compromises between true positive rate on in-distribution and false positive rate on out-of-distribution subsets of NWPU-RESISC45. As in [3], in both cases, thresholds are chosen so that the true positive rate is equal to 95%.

## IV. RESULTS

For both detector types the performances are noticeably better when the OOD dataset consists of non remote sensing imagery. Obviously, the distances between image embeddings of remote sensing OOD images and ID images are smaller than the distances between the image embeddings of remote sensing and non remote sensing images. This is especially pronounced in the case when the images from PatternNet are used as both ID and OOD images. In that case, due to the difference in image distributions between NWPU and PatternNet, it is harder to discern ID and OOD images.

Furthermore, from the results in Table I. we can see that the classifiers with higher ID accuracies result in better OOD detection performances in all the cases when MSP detector is concerned. This result is in line with the results in [11]. When a KNN detector is used, better ID classifiers result in better OOD detectors for Food5k and Imagenette OOD datasets. Tiny differences in classifier ID accuracies give significant improvements in OOD detecting performances on non remote sensing OOD datasets, which is not the case for remote sensing OOD datasets. In the experiments with NWPU-RESISC45 and PatternNet, however, the best KNN detector is obtained using embeddings from the classifier trained using only scheduler, which is not the best ID classifier. The most probable reason is again small distances between image embeddings of remote sensing ID images and OOD images, that make discrimination ID from OOD samples harder and disrupt the positive coupling between classifiers ID accuracy and OOD detector performance.

Table II. shows how the number of nearest neighbors, $k$, affects the KNN detector performance. We vary the number of neighbors k ∈ {1, 2, 5, 10, 20, 50, 100, 200} and we notice that AUROC value in each of the four experiments is almost independent of k. However, that is not the case for FPR values, whose changes are more significant, but different for each OOD set individually. To examine the effects of k on detector performance, we report the average FPR score over four OOD datasets. The results are shown in Fig. 1 and we can see that the average FPR values are the lowest and approximately equal for $k$ values between 1 and 50. For the values of $k$ greater than 50, the effect of the increase of $k$ on the detector performance is negative.

TABLE I. AUROC AND FPR VALUES FOR KNN AND MSP DETECTORS FOR DIFFERENT NETWORK TRAINING VARIANTS. ID ACCURACIES OF THE CLASSIFIERS ARE ALSO REPORTED. THE BEST RESULT FOR EACH OOD DATASET IS HIGHLIGHTED IN BOLD.

| Network training type/detector type | OOD dataset | | | | | | | | ID accuracy |
|---|---|---|---|---|---|---|---|---|---|
| | NWPU OOD | | Food5K | | Imagenette | | PatternNet OOD | | |
| | AUROC | FPR | AUROC | FPR | AUROC | FPR | AUROC | FPR | |
| Base/KNN | 90.11 | 41.06 | 98.45 | 7.64 | 94.16 | 25.63 | 84.05 | 40.95 | 91.64 |
| Base/MSP | 85.85 | 71.18 | 87.51 | 72.64 | 84.40 | 73.25 | 78.68 | 69.11 | |
| Augmentations/KNN | 92.54 | 34.40 | 99.05 | 3.92 | 96.45 | 18.65 | 83.37 | 38.98 | 92.01 |
| Augmentations/MSP | 88.27 | 63.45 | 86.22 | 72.92 | 84.54 | 75.18 | 79.39 | 68.25 | |
| Scheduler/KNN | **96.40** | **19.59** | 99.51 | 1.24 | 98.25 | 8.20 | **89.76** | **26.06** | 94.95 |
| Scheduler/MSP | 92.00 | 53.96 | 93.60 | 47.80 | 92.07 | 50.85 | 83.13 | 57.32 | |
| Augmentations and scheduler/KNN | 95.77 | 20.79 | **99.68** | **0.96** | **99.02** | **4.13** | 86.69 | 29.42 | **95.32** |
| Augmentations and scheduler/MSP | 91.26 | 49.55 | 92.70 | 46.56 | 91.37 | 49.07 | 83.66 | 49.01 | |

TABLE II. EFFECTS OF THE NUMBER OF NEAREST NEIGHBORS ON AUROC AND FPR.

| | OOD dataset | | | | | | | | | |
|---|---|---|---|---|---|---|---|---|---|---|
| k | NWPU OOD | | Food | | Imagenette | | PatternNet OOD | | Average | |
| | AUROC | FPR | AUROC | FPR | AUROC | FPR | AUROC | FPR | AUROC | FPR |
| 1 | 94.82 | 23.95 | 99.61 | 1.32 | 99.03 | 3.85 | 85.93 | 29.24 | 94.85 | 14.59 |
| 2 | 95.36 | 22.26 | 99.68 | 0.88 | 99.09 | 3.52 | 86.26 | 29.29 | 95.10 | 13.99 |
| 5 | 95.77 | 20.79 | 99.68 | 0.96 | 99.02 | 4.13 | 86.69 | 29.42 | 95.29 | 13.83 |
| 10 | 96.17 | 19.21 | 99.65 | 0.84 | 98.92 | 4.41 | 84.79 | 32.63 | 94.88 | 14.27 |
| 20 | 96.54 | 17.89 | 99.60 | 1.04 | 98.77 | 5.20 | 84.81 | 33.93 | 94.93 | 14.52 |
| 50 | 96.91 | 15.16 | 99.52 | 1.16 | 98.48 | 6.96 | 86.89 | 32.91 | 95.45 | 14.05 |
| 100 | 96.88 | 15.03 | 99.39 | 1.64 | 98.05 | 9.81 | 86.53 | 35.99 | 95.21 | 15.62 |
| 200 | 96.46 | 16.58 | 99.02 | 2.92 | 97.00 | 18.19 | 85.71 | 38.96 | 94.55 | 19.16 |

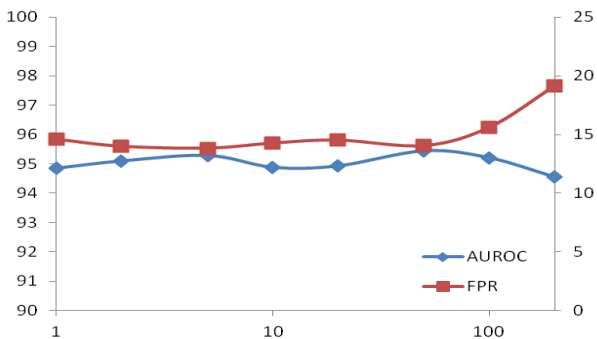

Figure 1. Effects of the number of nearest neighbors on the performance of the detector.

## V. CONCLUSION

In this paper we have proposed a benchmark for remote sensing OOD detection, and experimentally shown that the nearest neighbor OOD detector outperforms the MSP detector by a large margin. In addition, a KNN detector bring some other benefits. First, it is very intuitive to discriminate OOD from ID samples by measuring their distances from the training data contributing to the explainability of the detector. Furthermore, KNN method can be used with any type of classifier able to produce image embeddings, no matter how it was trained. Finally, it does not require OOD samples during training.

Although in our experiments KNN method outperformed MSP in OOD detection, it would be worthwhile to test some alternative distance based methods and potentially achieve better performance on the proposed benchmarks. For example, distance could be computed in higher-dimensional vector space in which the image embeddings would be concatenated with the predicted probabilities. Also, cross-entropy between the predicted class probability for the test image and the average value of predicted class probabilities for the $k$ nearest neighbors from the training set could be used as a distance measure. We plan to pursue these directions in the future work.

AKNOWLEDGMENT

This research was funded in part through the Provincial Secreteriat for Higher Education, Scientific and Research Activity, Autonomous Province of Vojvodina, Republic of Serbia, Project no. 142-451-1820/2022-01/1.